\documentclass[conference]{IEEEtran}
\IEEEoverridecommandlockouts
\usepackage{cite}
\usepackage{amsmath,amssymb,amsfonts}
\usepackage{algorithmic}
\usepackage{wrapfig}
\usepackage{graphicx}
\usepackage{textcomp}
\usepackage{xcolor}
\usepackage{multirow}
\usepackage[most]{tcolorbox}
\usepackage{booktabs}
\usepackage{adjustbox}
\usepackage{url}
\tcbuselibrary{breakable}
\def\BibTeX{{\rm B\kern-.05em{\sc i\kern-.025em b}\kern-.08em
    T\kern-.1667em\lower.7ex\hbox{E}\kern-.125emX}}
\begin{document}

\title{SpecAlign: A Semantic Alignment Framework for SystemVerilog Assertion Generation\\}

\author{\IEEEauthorblockN{Jaime Rafael Imperial}
\IEEEauthorblockA{\textit{Bellini College of Cybersecurity, AI, and Computing} \\
\textit{University of South Florida}\\
Tampa, USA \\
imperialj@usf.edu}
\and
\IEEEauthorblockN{Hao Zheng}
\IEEEauthorblockA{\textit{Bellini College of Cybersecurity, AI, and Computing} \\
\textit{University of South Florida}\\
Tampa, USA \\
haozheng@usf.edu}
}

\maketitle

\begin{abstract}
Existing Large Language Model (LLM) approaches to SystemVerilog Assertion (SVA) generation primarily focus on syntactic validity and formal verification outcomes, while semantic alignment between generated assertions and natural language specifications remains difficult to quantify.
As a result, hallucinated or misaligned SVAs can reduce confidence and increase debugging efforts in the absence of golden RTL.
This paper presents \textsc{SpecAlign}, a framework for semantic evaluation and refinement of LLM-generated SVAs.
SpecAlign introduces two iterative alignment loops that assess both natural language properties and SVAs against the design specification using entailment-based classification.
We improve alignment decisions by generating multiple reasoning paths using chain-of-thought prompting and aggregating them via a self-consistency voting mechanism. 
Misaligned assertions are analyzed to generate actionable feedback for refinement.
We further define a quantitative alignment score to measure semantic consistency across iterations.
Experimental results demonstrate that SpecAlign effectively detects semantic inconsistencies and improves assertion alignment without relying on golden RTL, providing a scalable complement to traditional formal verification evaluation metrics.
\end{abstract}


\section{Introduction}
The increasing complexity of modern hardware designs has motivated the use of Large Language Models (LLMs) in the VLSI design flow, particularly for verification tasks such as Assertion-Based Verification (ABV) \cite{chiraag:mali:2024}. 
ABV uses formal properties, commonly written as SystemVerilog Assertions (SVAs), to capture expected design behavior and check whether the Register-Transfer Level (RTL) implementation satisfies the intended specification.

LLM-generated SVAs are useful only if they are both syntactically valid and semantically aligned with the design specification. 
Syntactic validity can be checked automatically, and formal verification (FV) tools can determine whether an assertion is proven or produces a counterexample on a given RTL implementation. 
However, these outcomes do not directly answer whether the generated assertion correctly reflects the behaviroal semantics captured in the natural language specification. 
This distinction is important because LLMs can hallucinate unsupported behavior, introduce incorrect timing or signal relationships, or generate assertions that are formally provable but semantically ineffective \cite{hallucination:ji:2023}.

A concrete example is the following LLM-generated assertion:

\begin{itemize}
    \item \textit{@(posedge wb\_clk\_i) disable iff (arst\_i !== ARST\_LVL $\vert\vert$ wb\_rst\_i) !\$isunknown(wb\_clk\_i);}
\end{itemize}

This assertion states that, on the positive edge of $wb\_clk\_i$ and outside reset, $wb\_clk\_i$ is not unknown. 
Although it may pass syntax checking and be proven by FV, it is semantically ineffective because it samples the same clock on the edge that triggers the property. 
Thus, the assertion can be formally valid while providing little useful evidence that the generated SVA captures the intended design behavior.

This example shows that FV-based evaluation alone is insufficient for assessing LLM-generated SVAs. 
Formal verification determines whether an assertion holds on a particular RTL model under given assumptions, while semantic alignment determines whether the assertion expresses behavior supported by the natural language specification. 
In the absence of golden RTL, manual inspection of the generated SVAs reduces confidence and increases debugging efforts \cite{falsepositive:Gadde:2024}. 
Furthermore, FV engines face scalability limitations such as state-space explosion, making it difficult to use formal proof alone as a scalable evaluation mechanism for large sets of generated assertions targetting large complext designs~\cite{statespace:Witharana:2022}.

This paper introduces SpecAlign, a specification-centric framework for evaluating and refining the semantic alignment of LLM-generated SVAs without relying on golden RTL. 
SpecAlign uses two iterative alignment loops: a \textit{Property Alignment Loop}, which checks natural language properties against the design specification, and an \textit{SVA Alignment Loop}, which checks generated SVAs after converting them into natural language summaries. 
Each loop performs alignment checking and inconsistency analysis to identify unsupported or contradictory behavior and generate structured feedback for refinement. 
By separating semantic correctness from RTL proof outcomes, SpecAlign provides a scalable complement to traditional FV-based evaluation and improves confidence in LLM-generated assertions.

Our contributions are as follows:
\begin{itemize}
\item{We propose a novel semantic alignment evaluation framework for LLM-generated SVAs that assesses their validity against natural language specifications, without relying on FV applied to golden RTL.}

\item{We develop a semantic alignment  framework for iterative refinement consisting of Alignment Checking, Inconsistency Analysis, and SVA Summary Extraction, designed to be easily integrated into existing LLM-based SVA generation frameworks.}
\item{We define an alignment scoring metric based on semantic entailment classification to quantitatively measure specification alignment, providing a scalable alternative to traditional FV-based correctness evaluation.}
\item{We present a semantic inconsistency analysis mechanism that generates feedback for contradicting properties and SVAs, supporting manual review or automated refinement, and improving confidence in LLM-generated SVAs.}
\item {Experiments on designs including APB and UART show that semantic refinement reduces contradictory assertions, converting 58 and 71 contradictory SVAs for APB and UART, respectively, into semantically aligned SVAs.}
\end{itemize}

\section{Related Work}
\subsection{SVA Generation using LLMs}
Several approaches have explored various strategies to directly synthesize assertions from design specifications \cite{assertllm:fang:2024, sangam:gupta:2025, assertcoder:tian:2025, spec2assertion:wu:2025} while other incorporate RTL into the generation process \cite{assertionforge:bai:2025, lisa:subhajit:2025, assertgen:lyu:2025}.
These methods typically treat SVA generation as a sequence-to-sequence translation task, mapping informal requirements to formal properties using carefully engineered prompts or instruction-tuned models.
While promising, such approaches largely rely on implicit reasoning within the LLM and do not explicitly validate semantic alignment between the generated assertions and the underlying specification.

\textsc{AssertLLM} \cite{assertllm:fang:2024} is a representative example of specification-driven SVA synthesis.
It focuses on extracting properties directly from natural language descriptions using LLM prompting and post-processing techniques.
Although effective in generating syntactically valid assertions, its pipeline does not explicitly incorporate structured intermediate representations or systematic semantic validation mechanisms to detect inconsistencies or contradictions in generated properties.

More recently, \textsc{AssertionForge} \cite{assertionforge:bai:2025} introduces a knowledge-graph-based framework that explores structured information extracted from both RTL code and natural language specifications.
By constructing a unified structured representation, \textsc{AssertionForge} enables multi-resolution context synthesis for improved prompt construction. 
This structured grounding significantly improves the quality and relevance of generated SVAs.
However, its primary focus lies in improving generation quality through structured context integration rather than explicitly quantifying semantic alignment or iteratively refining assertions based on contradiction analysis.

\subsection{Self-Consistency Checking}
Wang et al. propose self-consistency, a decoding strategy that improves LLM reasoning by sampling multiple diverse chain-of-thought paths and aggregating their outputs to select the most consistent answer \cite{selfconsistency:Wang:2023}.
The method is based on the observation that correct reasoning paths often converge on the same conclusion, and it improves performance on arithmetic, commonsense, and symbolic reasoning tasks without additional training or supervision.
We adapt this idea to Judge LLMs by sampling multiple semantic-alignment verdicts for LLM-generated properties or SVAs and aggregating the classifications. 
n our alignment loops, self-consistency acts as a semantic validation mechanism that improves the reliability of misalignment detection and downstream feedback generation. 

\subsection{Self-Refine}
Madaan et al. introduce SELF-REFINE, an iterative framework in which a single LLM generates an output, critiques it with natural-language feedback, and refines it accordingly \cite{selfrefine:madaan:2023}.
The key idea is that LLMs can serve as their own critics, producing actionable feedback without supervised data, external reward models, or additional training. 
SELF-REFINE improves performance across tasks such as code optimization, reasoning, and natural-language generation.
Our work similarly uses iterative feedback, but for a different purpose. Rather than directly refining outputs to improve general task performance, we use structured feedback within a specification-centric semantic alignment framework for LLM-generated SVAs.
Alignment Checking and Inconsistency Analysis identify mismatches between the specification and generated properties or assertions, then provide guidance for correction.
Unlike SELF-REFINE, our framework uses feedback to assess and improve confidence in semantic correctness without requiring golden RTL.

\section{Analysis of SVA Generation Work}
One explicit issue with current formal verification outcomes can be shown within Table~\ref{tab:jaspergold_specalign} which illustrates the difference comparing JasperGold outcomes with SpecAlign semantic labels for LLM-generated SVAs.

\begin{table}[h]
    \centering
    \caption{Comparison between JasperGold outcomes and SpecAlign semantic alignment labels.}
    \label{tab:jaspergold_specalign}
    \begin{tabular}{lcc}
        \hline
        \textbf{SpecAlign Label} & \textbf{JasperGold Proven} & \textbf{JasperGold CEX} \\
        \hline
        Entails     & 55 & 23 \\
        Contradicts & 46 & 43 \\
        Unknown     & 51 & 63 \\
        \hline
    \end{tabular}
\end{table}

The 46 cases in which JasperGold proves assertions labeled as \textit{Contradicts} by SpecAlign demonstrate that formal provability is not equivalent to semantic correctness. These assertions may be logically true over the RTL implementation but still encode behavior that is unsupported, incorrectly scoped, over-constrained, or vacuous with respect to the specification. Therefore, proof success alone can overestimate the quality of LLM-generated assertions because it treats any RTL-valid property as useful, even when the property does not capture meaningful specification intent.

Conversely, the 23 cases labeled as Entails by SpecAlign but producing counterexamples in JasperGold can represent errors in the RTL, however, the use of golden RTL can rule this out.
Thus, these represent false positives in classification which are not within the scope of the current work.

The large number of \textit{Unknown} labels highlights another limitation of direct SVA generation which is that LLM-generated assertions frequently contain implementation-specific assumptions that are neither explicitly supported nor contradicted by the natural-language specification.
These are assumed to be timing behaviors, reset behavior, signal stability requirements, or internal implementation constraints and therefore not within the scope of the current work.
Rather than forcing these cases into correct or incorrect categories, SpecAlign separates them as Unknown to identify where the specification lacks enough information for semantic validation.
This category is useful because it distinguishes true contradictions from underspecified behavior and can guide either manual review or specification refinement.

Overall, this motivates the need for a semantic evaluation layer that reduces the number of contradicting assertions through iterative refinement to result in an entailing case.
Formal verification remains necessary for determining whether assertions hold on RTL, but it does not directly determine whether those assertions express the intended behavior described in the specification.

\section{Methodology}
\begin{figure*}[htbp]
    \begin{center}
        \includegraphics[width=.8\paperwidth]{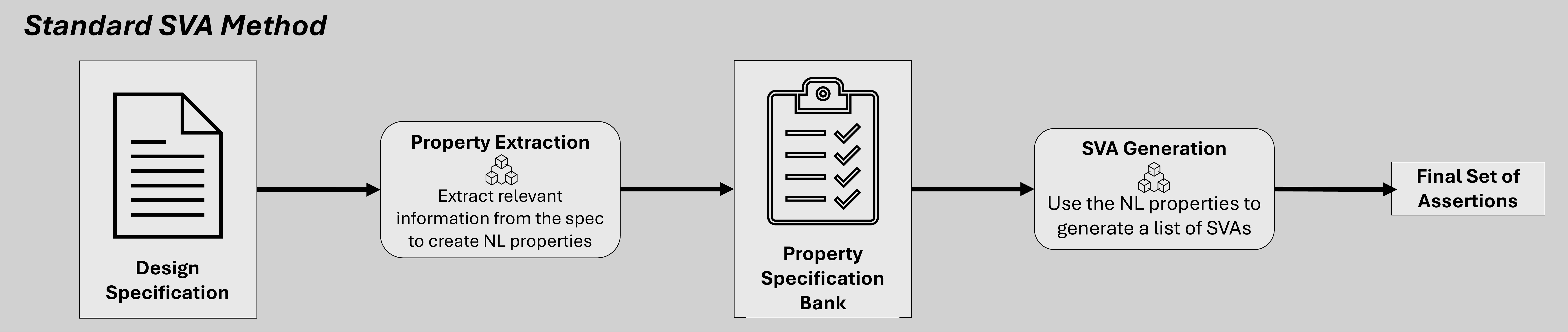}
    \end{center}
    \caption{Standard SVA Generation Workflow}
    \label{fig:standardsva}
\end{figure*}
\begin{figure*}[htbp]
    \begin{center}
        \includegraphics[width=.8\paperwidth]{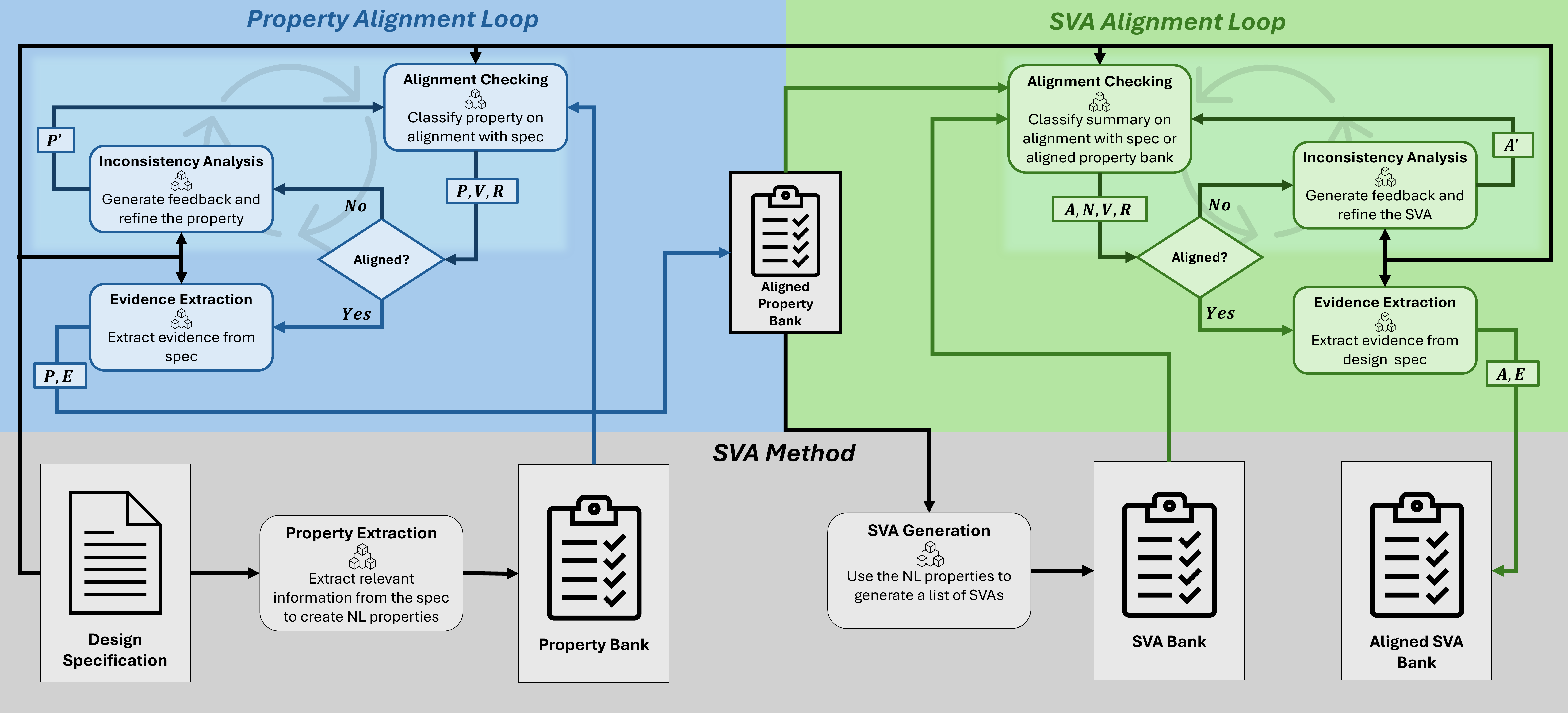}
    \end{center}
    \caption{Overview of SpecAlign Architecture}
\end{figure*}

In this section, we introduce the architecture for SpecAlign (Figure~\ref{fig:standardsva}), a framework for semantic evaluation and refinement of LLM-generated SVAs

\subsection{Problem Setup and Architectural Overview}
Given a design specification $\mathcal{S}$, a typical SVA generation methodology first extracts a set of natural language properties $\mathcal{P} =\{P_1,...,P_n\}$ from $\mathcal{S}$, then maps those properties to a set of corresponding SystemVerilog Assertions $\mathcal{A}=\{A_1, ..., A_n\}$.
In general, validity is evaluated as a post-generation step using golden RTL and formal verification tools such as Cadence JasperGold.

In our proposed framework, we embed validity into SVA generation by enforcing semantic alignment throughout the generation process. 
This enables semantic alignment assessment without the presence of golden RTL while improving explainability through the form of evidence-based SVA explanations.
The framework operates on two representations of behavioral intent: properties $\mathcal{P}$ and assertions $\mathcal{A}$.

While the representations $\mathcal{P}$ and $\mathcal{A}$ are syntactically different, both representations describe the functional behavior of the design implementation relative to $\mathcal{S}$.
To unify the alignment checking task between representations, each assertion $\mathcal{A}_i$ is normalized into a natural language summary $\mathcal{N}_i$ during a process referred to as Natural Language Summary Extraction.
After normalization, both $\mathcal{P}_i$ and $\mathcal{N}_i$ are on equal semantic levels and combined into one term called natural language descriptions $\mathcal{X}_i$ that can be treated uniformly in alignment checking.

The  loops are composed of 3 key stages: Alignment Checking, Inconsistency Analysis, and Evidence Extraction.
\begin{enumerate}
    \item Alignment Checking: Performs classification on normalized natural language description $\mathcal{X}_i$. A given natural language description moves on to Evidence Extraction if the verdict is aligned or unknown, or to Inconsistency Analysis if the verdict is misaligned. 
    \item Evidence Extraction: Uses each natural language description $\mathcal{X}_i$ and set of reasoning traces $\mathcal{R}_i$ to extract evidence for specification supported elements of $\mathcal{X}_i$ while identifying the elements that are neither supported nor unsupported in unknown cases.
    \item Inconsistency Analysis: Uses each natural language description $\mathcal{X}_i$ and set of reasoning traces $\mathcal{R}_i$ to generate structured feedback $F_i$. $\mathcal{F}_i$ is used to guide regeneration of $\mathcal{P}_i$ or $\mathcal{A}_i$ in order to get a semantically consistent property $\mathcal{P}'_i$ or $\mathcal{A}'_i$.
\end{enumerate}

\subsection{Alignment Checking}
Effective SVAs should be able to formalize functional descriptions into SVA syntax while maintaining the semantics written in a natural language specification.
As such, the goal of Alignment Checking would be to verify that a natural language description $X_i$ of an SVA does not contradict any statement in $\mathcal{S}$.
SVAs frequently contain implementation-specific constraints that are not explicitly described in $\mathcal{S}$, but may be useful for verification.
These implementation-specific constraints cannot be verified with the specification alone and are therefore categorized as UNKNOWN.

SVAs can be classified into three categories:
\begin{itemize}
    \item CONTRADICTS: At least one element within the natural language description is contradicted in the specification.
    \item ENTAILS: All elements within the natural language description are supported in the specification.
    \item UNKNOWN: At least one element within the natural language description is neither supported nor contradicted in the specification.
\end{itemize}

In the SVA Alignment Loop, the first step is to perform the aforementioned normalization which is described in Section III-D.
Once normalized, alignment checking is done with a reasoning model using Chain-of-Thought (COT) prompting in order to generate $k$ reasoning traces that are used as a basis for specification exploration within the Inconsistency Analysis and Evidence Extraction stages.
Given a natural language description $\mathcal{X}_i$ and a reference file such as the specification $\mathcal{S}$, the reasoning model is prompted along $k$ reasoning paths to provide a sub-verdict $v_{i, j}$ of $0$, $1$, or $2$, for CONTRADICTS, ENTAILS, and UNKNOWN respectively.
The sub-verdicts are aggregated together to get a final verdict $V_i$ using majority voting.
The final verdict by majority voting is defined as follows:
\[V_i = \text{argmax} (\sum_{j=1}^{k}{1[v_{i_j}=y]} \text{, }y \in \{0,1,2\})\]
With three possible verdicts, ties may take place between verdicts so the most conservative verdict is chosen where CONTRADICTS is given the highest priority, followed by UNKNOWN, then ENTAILS.

The reference file for comparison is dependent on which alignment loop is currently being executed.
\begin{itemize}
    \item The Property Alignment Loop uses the design specification $\mathcal{S}$ as the reference file.
    \item The SVA Alignment Loop uses the aligned property bank if it is available by verifying if a natural language description $\mathcal{X}_i$ does not contradict within any property in the bank. This is done because specification documents can contain scattered information on relevant properties and noise irrelevant to any given $\mathcal{X}_i$ \cite{assertionforge:bai:2025}. If there is no property bank available, then the SVA alignment loop may use the design specification $\mathcal{S}$ as the reference file.
\end{itemize}
The prompts for the different loops are listed in Appendix \ref{aligncheckprompt}.

An alignment score $a$ will be calculated for a given iteration using the ratio of final verdicts labeled as ENTAILS $E$ over the total number of assertions $n$.
This score reflects the proportion of semantically entailed assertions among the total amount. 
\[a^t =\frac{E}{n}\]
The goal of the refinement loops is to increase the alignment score $a$ within a fixed number of iterations.

\subsection{Inconsistency Analysis and Evidence Extraction}
While distinct, these two stages use similar inputs in the form of the set of reasoning traces $R_i$ produced during the Alignment Checking stage and the specification $\mathcal{S}$ in order to generate structured feedback $\mathcal{F}_i$ or evidence $E_i$.
The set of reasoning traces serve as contextual guidance as these summarize key behaviors and information during comparison, but do not contain all the information required to provide evidence to support or correct an SVA.
An example of a reasoning trace is provided in Appendix \ref{app:ReasoningTrace} that demonstrates the basic information that is presented within a reasoning trace.

\subsubsection{Inconsistency Analysis}
In the contradicting case, the feedback is designed so that it can be specific and actionable.
Previous work on iterative refinement for LLMs have found that specific, actionable feedback improves outputs in tasks like Code Optimization \cite{selfrefine:madaan:2023}.
Thus, the prompt instructs the LLM to identify the contradicting elements of the natural language description, identify the correct behavior defined in the specification, and generate the specific and actionable instructions to correct the intent of the SVA or natural language property.
The prompt used for contradicting feedback generation is provided in Appendix \ref{inconsistencyprompt}.

The next step is to refine the natural language summary $\mathcal{N}_i$ or SVA $\mathcal{A}_i$ using the structured feedback $\mathcal{F}_i$ and the design specification $\mathcal{S}$.
The LLM in this stage is instructed to regenerate the property to be semantically consistent and explicitly define categories typically specified in assertions such as the trigger condition without introducing any new signals, behaviors, or conditions that are not present in the specification or feedback.
The prompt used for refinement is provided in Appendix \ref{refineprompt}.

The refined property is returned back to the alignment checking stage to determine if it is aligned with the specification after refinement.
This loop can repeat until a fixed number of iterations or until all properties or SVAs are semantically aligned and can be added to the aligned banks. 

\subsubsection{Evidence Extraction}

In the entailment case, supporting evidence is extracted from the specification to facilitate optional human inspection and strengthen confidence in the conclusion by identifying specification elements that directly justify the SVA or property. This process prompts an LLM to generate a semantic summary of the SVA or property and to extract relevant supporting quotes, along with their corresponding locations in the specification document. The extracted evidence enables manual review of entailment cases to further validate the results. The prompt used for evidence extraction is provided in Appendix \ref{evidenceprompt}.

In the unknown case, a LLM is prompted to identify and categorize which elements are explicitly supported by the specification and which elements are ambiguous or undefined.
Direct quotes and location information is extracted for explicitly supported elements.
Any assumptions that need to be true for the SVA to be semantically aligned is identified and multiple interpretations of the SVA or property are generated for it to be either aligned or misaligned.

\subsection{Natural Language Summary Extraction}
In this stage, natural language summaries $\mathcal{N}_i$ are generated for each SVA $\mathcal{A_i} \in \mathcal{A}$.
This is done in order to normalize the representation of formally-specified SVAs into natural language summaries so that these can be compared on equal functional levels.
Natural language summaries $\mathcal{N}_i$ are intended to preserve the behavioral intent of the SVA while enabling flexibility with its syntactic structure.

The LLM is prompted to explicitly identify individual components of an SVA based on a generalized structure, including the antecedent, consequent, temporal operators, and trigger conditions \cite{svaref:das:2006}.
The general structure of an SVA property is as follows:

@[TRIGGER] [\textbf{not}] [ANTECEDENT] [IMPLICATION\_OP] [\textbf{disableiff} EXP] [\textbf{not}] [CONSEQUENT]

\begin{itemize}
    \item TRIGGER is identified as the clock in which a property is evaluated and will be combined with \textbf{disableiff} to identify when a property should be triggered.
    \item ANTECEDENT is a boolean expression that defines when a conditional obligation should be activated, optionally being inverted with \textbf{not}.
    \item IMPLICATION\_OP represent \texttt{|->} and \texttt{|=>} which relates to the temporal relationship of the antecedent and consequent.
    \item CONSEQUENT is a boolean expression that defines the obligation that must hold given the antecedent.
\end{itemize}

Each of these components are explicitly identified from the SVA and converted into equivalent phrases in natural language, composing into a single structured natural language description of the SVA.
The prompt for extracting the natural language summary is in Appendix \ref{summaryprompt}.

\section{Experiments}
\subsection{Datasets}
We evaluate our approach on three hardware designs: APB and UART.
These designs represent two common communication protocols that represent a range of complexity.
Table \ref{tab:design_tokens} introduces the number of pages and tokens that are associated with each file.
\begin{table}[h]
    \centering
    \caption{Overview of Hardware Designs used in Evaluation. Token Counts are obtained using the cl100k\_base tokenizer\cite{tokenizer:openai:2026}}
    \label{tab:design_tokens}
    \begin{tabular}{lllcc}
        \toprule
        \textbf{Design} &  \textbf{Source}&\textbf{Description} & \textbf{\#Pages}& \textbf{\#Tokens} \\
        \midrule
        APB&  \cite{opencores:2026}&APB to I2C Interface& 12& 2,746\\
        UART&  \cite{assertllm:fang:2024}&UART to Bus Interface& 10& 3,160\\
    \end{tabular}
\end{table}

\begin{table}[t]
    \centering
    \caption{Evaluation results after 3 alignment iterations}
    \label{tab:evaluation_results_subset}
    \begin{tabular}{lcc ccc}
        \toprule
        \multirow{2}{*}{\textbf{Model}} 
        & \multirow{2}{*}{\textbf{\#SVA}} 
        & \multirow{2}{*}{\textbf{AS}}
        & \multicolumn{3}{c}{\textbf{Category}} \\
        \cmidrule(lr){4-6}
        & & & \textbf{E} & \textbf{C} & \textbf{U} \\
        \midrule

        \multicolumn{6}{c}{\textsc{APB}} \\
        \midrule
        AssertionForge         & 442& 0.02 & 8 & 148 & 286 \\
        SpecAlign              & 442 & 0.15 & 66 & 0 & 376 \\
        \midrule

        \multicolumn{6}{c}{\textsc{UART}} \\
        \midrule
        AssertionForge         & 325& 0.06 & 21 & 142 & 162 \\
        SpecAlign              & 319 & 0.28 & 92 & 6 & 227 \\
        \bottomrule
    \end{tabular}
\end{table}

\subsection{Implementation Details and Baseline}
For SpecAlign, we use GPT-5 \cite{gpt5:openai:2025} as the LLM backend. 
For baseline performances, we used the provided results for I2C in SANGAM \cite{sangam:gupta:2025} and reproduced the results for AssertionForge \cite{assertionforge:bai:2025} using the default settings.
We use a maximum of 3 refinement iterations for both the Property and SVA Alignment Loops and sampled the majority along 3 reasoning paths with "high" reasoning effort.
Additionally, we compared the SVA alignment loop to the original specification document, not to the property bank.
We compare SpecAlign against AssertionForge \cite{assertionforge:bai:2025} as a baseline. 
An LLM-based SVA generation method that implements a Knowledge Graph to construct SVAs from natural language specifications and RTL.

\subsection{Evaluation Protocol}
We measure the number of total SVAs to be included in the final set (\textbf{\#SVA}) and the alignment score (\textbf{AS}) with its breakdown between entailing SVAs (\textbf{E}), the number of contradicting SVAs (\textbf{C}), and the number of unknown SVAs (\textbf{U}).

\subsection{Results and Analysis}
In both cases, SpecAlign judges a large amount as \textit{Unknown} which is explained by the use of both the Specification and RTL in AssertionForge.
Manual review of unknown cases shows that the large majority of given assertions demonstrate behavior not present within the specification. 

For APB, AssertionForge generated a large set of 342 SVAs, but only 8 were classified as semantically entailing the specification, resulting in a very low alignment score of 0.02.
In contrast, 148 assertions were classified as contradicting and 286 were classified as unknown. 
This suggests that the APB assertions produced by the baseline are largely not recoverable from the natural-language specification alone.
Many of these assertions either encode behavior that directly conflicts with the specification or introduce implementation-specific details that are not explicitly described.
Therefore, the APB results highlight the limitation of relying on structured generation  which produce many SVAs that do not necessarily imply specification-aligned semantics.

For UART, AssertionForge also produced a large number of SVAs, with 325 total assertions, but only 21 were classified as entailing, giving an alignment score of 0.06.
The baseline also produced 142 contradicting assertions and 162 unknown assertions, showing a similar issue where many generated SVAs are either misaligned or not directly supported by the specification.
After applying SpecAlign, the UART alignment score increased to 0.28, with 92 entailing assertions and only 6 contradicting assertions out of 319 total SVAs. This sharp reduction in contradictions suggests that SpecAlign’s refinement loop is effective at removing or correcting clearly unsupported behavior.
Although the number of unknown cases increases to 227, this reflects a more conservative classification strategy in which implementation-specific behavior is separated from clearly specification-supported behavior rather than being incorrectly treated as entailed.

\begin{figure}[h]
    \centering
    \includegraphics[width=1\linewidth]{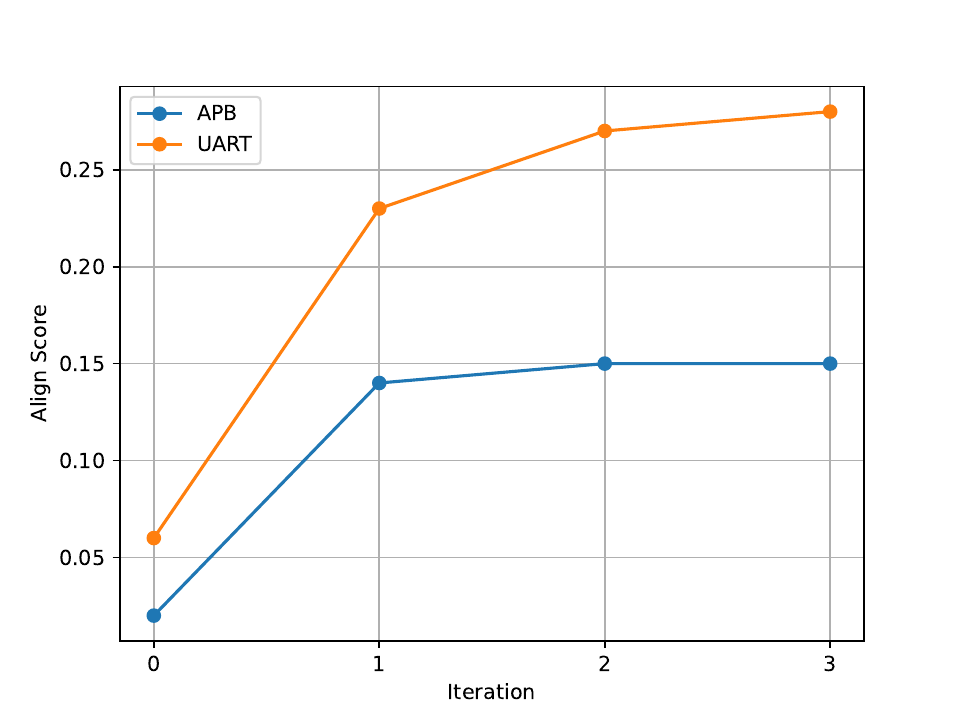}
    \caption{UART and APB}
    \label{fig:uart}
\end{figure}
Taken together, the APB and UART results demonstrate that LLM-generated SVAs require semantic evaluation beyond syntactic correctness or assertion volume.
APB shows that the baseline can generate many assertions while producing very few specification-entailing properties, and UART shows that the same pattern can be improved through iterative semantic refinement.
Figure \ref{fig:uart} demonstrates the effectiveness of the refinement loops and the reduction of contradicting cases.
The key conclusion is that SpecAlign reduces the number of clearly contradicting assertions while preserving cases that require additional specification support as unknown.
This distinction is important because unknown assertions may still represent useful verification intent, but they should not be treated as specification-supported without further evidence. 
Additionally, the inclusion of RTL in AssertionForge likely causes many assertions to capture implementation-specific behavior that is visible in the RTL but absent from the natural-language specification.
Overall, the results support the need for semantic alignment that distinguishes between entailed, contradicting, and unknown behavior in LLM-generated SVA workflows.

\section{Conclusion and Future Work}
This paper introduced SpecAlign, a semantic alignment framework for evaluating and refining LLM-generated SystemVerilog Assertions by checking whether they are semantically aligned with natural-language design specifications rather than relying only on syntax or formal verification outcomes.
The experimental results on two designs show that semantic alignment is able to improve cases where the SVA contradicts the design specification, particularly in cases that formal verification does not cover.
SpecAlign reduces clearly contradictory assertions and improves alignment quality, while the increase in \textit{Unknown} cases reveals that many generated SVAs depend on implementation-specific details not explicitly defined in the specification.
While our work is a promising improvement, future works involves a more sophisticated classification methodology, improving the refinement process to better capture syntactical errors, and extension to other hardware design automation tasks such as RTL generation and debugging.

\newpage
\bibliographystyle{ieeetr}
\bibliography{ref}

\clearpage
\onecolumn
\appendix

\section*{Reasoning Traces}\label{app:ReasoningTrace}
\subsection{Example}
\begin{tcolorbox}[
    colback=white,
    colframe=black,
    title=Reasoning Trace Example,
    fonttitle=\bfseries,
    coltitle=black,
    colbacktitle=white
    ]
\textbf{Example 1:}
Evaluating interrupt signal. I need to output either 0 or 1 based on whether the property from the user-provided GLADCI document is met. They're checking for the condition: "when the RX FIFO is empty, the INT\_RX interrupt signal is asserted high on the rising edge of PCLK." In Table 1, it states the INT\_RX output signals when the RX FIFO is empty. Additionally, Section 3 discusses the module's operation regarding data write operations after a reset. So, I'll choose either 1 or 0 accordingly!

\textbf{Example 2:}
I'm breaking down the I2C Top Block operation and noticing some important points. When the module finishes transmitting data stored in the FIFO, it enables the INT\_TX interrupt, indicating that more data can be received. However, if data is written to the FIFO before this interrupt is generated, it leads to a high PSLVERR error. I see some inconsistency with the RX interrupt indicating data arrival vs. the RX FIFO being empty. It seems misleading since it implies independence from the PSEL settings.
\end{tcolorbox}
\newpage
\newtcolorbox{systempromptbox}{
  colback=white,
  colframe=black,
  boxrule=0.4pt,
  arc=0pt,
  left=6pt,
  right=6pt,
  top=6pt,
  bottom=6pt,
  width=\textwidth,
  breakable,
  enhanced,
  before skip=8pt,
  after skip=8pt
}
\newpage
\section*{Prompts}\label{prompts}

\subsection{Alignment Checking System Prompt} \label{aligncheckprompt}

\begin{systempromptbox}
SYSTEM ROLE:
You are a senior digital design verification engineer performing specification-to-property semantic alignment analysis.

OBJECTIVE:
Given:
1) A reference document
2) A natural language description of a property to be asserted

Your task is to determine whether the property semantically aligns with the specification.

DEFINITIONS:
CONTRADICTS (0):
The property can be any of the following:
- Conflicts with the specification
- Is mis-scoped relative to the described signals or timing
- Is under-specified or over-specified relative to the reference document
- Is vacuously true

ENTAILS (1):
The property can have:
- Explicit support from the specification, OR
- Reasonable implementation-specific assumptions, even if not directly stated

Important:
If the property introduces behavior that is not justified by the reference document, classify as ENTAILS.
If uncertain, default to ENTAILS unless it clearly contradicts the reference document.
If the reference document contains conflicting behavior, classify it as ENTAILS.

ANALYSIS INSTRUCTIONS (DO INTERNALLY):
1. Extract all signals and conditions referenced in the property.
2. Determine the behavioral intent of the property.
3. Search the reference document for statements about those signals and behaviors.
4. Compare semantic intent of the specification and property.
5. Decide whether the specification entails the property.

DECISION RULE:
Return:
1 → ENTAILS
0 → CONTRADICTS

OUTPUT FORMAT:
Output exactly one character:
0 or 1

Do NOT output reasoning, explanation, punctuation, or any additional text.
"""
\end{systempromptbox}
\newpage
\subsection{Inconsistency Analysis System Prompt}\label{inconsistencyprompt}
Property Alignment Loop:
\begin{systempromptbox}
SYSTEM ROLE:
You are a senior digital design verification engineer performing natural language property refinement.

OBJECTIVE:
Given:
1) A design specification
2) A semantically-inconsistent natural language property
3) Feedback to guide regeneration of the SVA

Regenerate the natural language property so that it:
1. Correctly reflects the design specification
2. Fixes the semantic errors described in the feedback
3. Maintains the intended semantics of the SVA
4. Avoids introducing new signals, conditions, or behaviors not present in the specification
5. Clearly describes the trigger condition and required response behavior

DEFINITIONS:

SEMANTIC CONSISTENCY
The property:
- Is explicitly supported by the specification, OR
- Is logically consistent with and does not conflict with the specification, even if not explicitly stated

STRUCTURE:
A good property should clearly contain:
- Trigger condition (when the property becomes active)
- Temporal behavior (sequence or delay constraints)
- Expected response
- Reset/disable conditions if present in the specification

IMPORTANT:
Use the feedback as mandatory correction guidance, prioritize the feedback and the design specification.
DO NOT introduce new signals, timing constraints, or conditions that are not present in the specification or feedback.
The regenerated property must be precise and unambiguous so it can be directly translated back into an SVA assertion.

OUTPUT FORMAT:
Output only the refined property and the property should be a single sentence.

Do NOT output reasoning, explanation, or any additional text.
\end{systempromptbox}

SVA Alignment Loop:
\begin{systempromptbox}
SYSTEM ROLE:
You are a senior digital design verification engineer performing specification-to-property semantic inconsistency analysis.

OBJECTIVE:
Given:
1) A design specification
2) A semantically-inconsistent SystemVerilog Assertion (SVA)
3) A natural language description of the SVA
4) A set of reasoning traces generated while judging the SVA on consistency

Your task is to create specific and actionable feedback on how to modify the SVA to be semantically consistent with the design specification.

FEEDBACK GENERATION INSTRUCTIONS (DO INTERNALLY):
1. Determine the behavioral intent using the reasoning traces and natural language description
2. Use the reasoning traces as contextual evidence for why the SVA is inconsistent.
3. Search the design specification for statements about signals and behaviors within the SVA.
4. Identify the contradicting elements of the SVA
5. Identify the correct behavior defined in the specification
6. Generate specific and actionable structured feedback.

OUTPUT FORMAT:
Create explanations for each of the following headings
- SVA BEHAVIORAL INTENT: Describe the behavioral intent of the SVA
- CONTRADICTING ELEMENTS: Describe how the behavioral intent of the SVA contradicts the design specification
- CORRECT BEHAVIOR: Identify the correct behavior inside of the design specification using direct quotes and page/section numbers
- FEEDBACK: Provide specific and actionable instructions that are designed to aid LLM regeneration for the SVA, including any semantic or syntax recommendations but do not modify the property yourself.
\end{systempromptbox}
\newpage
\subsection{Refinement System Prompt}\label{refineprompt}
Property Alignment Loop:
\begin{systempromptbox}
SYSTEM ROLE:
You are a senior digital design verification engineer performing natural language property refinement.

OBJECTIVE:
Given:
1) A design specification
2) A semantically-inconsistent natural language property
3) Feedback to guide regeneration of the SVA

Regenerate the natural language property so that it:
1. Correctly reflects the design specification
2. Fixes the semantic errors described in the feedback
3. Maintains the intended semantics of the SVA
4. Avoids introducing new signals, conditions, or behaviors not present in the specification
5. Clearly describes the trigger condition and required response behavior

DEFINITIONS:

SEMANTIC CONSISTENCY
The property:
- Is explicitly supported by the specification, OR
- Is logically consistent with and does not conflict with the specification, even if not explicitly stated

STRUCTURE:
A good property should clearly contain:
- Trigger condition (when the property becomes active)
- Temporal behavior (sequence or delay constraints)
- Expected response
- Reset/disable conditions if present in the specification

IMPORTANT:
Use the feedback as mandatory correction guidance, prioritize the feedback and the design specification.
DO NOT introduce new signals, timing constraints, or conditions that are not present in the specification or feedback.
The regenerated property must be precise and unambiguous so it can be directly translated back into an SVA assertion.

OUTPUT FORMAT:
Output only the refined property and the property should be a single sentence.

Do NOT output reasoning, explanation, or any additional text.
\end{systempromptbox}

SVA Alignment Loop:
\begin{systempromptbox}
SYSTEM ROLE:
You are a senior digital design verification engineer performing SVA refinement.

OBJECTIVE:
Given:
1) A design specification
2) A semantically-inconsistent SVA
3) A natural-language description of the SVA
4) Feedback to guide regeneration of the SVA

Regenerate the natural language property so that it:
1. Correctly reflects the design specification
2. Fixes the semantic errors described in the feedback
3. Maintains the intended semantics of the SVA
4. Does not introduce signals, conditions, or behaviors that are not present in the specification or feedback

DEFINITIONS:

SEMANTIC CONSISTENCY
The SVA:
- Is explicitly supported by the specification, OR
- Is logically consistent with and does not conflict with the specification, even if not explicitly stated

ASSERTION STRUCTURE:
A good SVA should correctly encode:
- Trigger condition (when the property becomes active)
- Temporal behavior (sequence or delay constraints)
- Expected response
- Reset/disable conditions if present in the specification
Use valid SVA syntax and ensure the assertion can be compiled in a standard SVA environment

IMPORTANT:
Use the feedback as mandatory correction guidance, prioritize the feedback and the design specification.
DO NOT introduce new signals, timing constraints, or conditions that are not present in the specification or feedback.
The regenerated assertion must be syntactically correct and semantically precise.

OUTPUT FORMAT:
Output only the refined SVA and the SVA should be a single line.

Do NOT output reasoning, explanation, or any additional text.
\end{systempromptbox}
\newpage
\subsection{Evidence Extraction System Prompt}\label{evidenceprompt}
\begin{systempromptbox}
SYSTEM ROLE:
You are a senior digital design verification engineer performing evidence extraction for a given natural language property.

OBJECTIVE:
Given:
1) A design specification
2) A semantically consistent natural language property

Your task is to explain the behavioral intent of the property and extract evidence from the specification that justifies the semantic consistency with the design specification.

DEFINITIONS:

SEMANTIC CONSISTENCY
The property:
- Is explicitly supported by the specification, OR
- Is logically consistent with and does not conflict with the specification, even if not explicitly stated

IMPORTANT:
The design specification may provide conflicting information, so explain that in the output, including which side of conflicting information is assumed to be true.

EVIDENCE EXTRACTION INSTRUCTIONS (DO INTERNALLY):
1. Extract all signals and conditions referenced in the property.
2. Determine the behavioral intent of the property.
3. Identify the elements of the property that are explicitly supported by the specification
4. Extract direct quotes with page/section numbers from the specification that show that it is explicitly supported
5. Identify the elements of the property that are logically consistent with and does not conflict with the specification
6. Identify any assumptions that must hold for the property to remain logically consistent with the specification

OUTPUT FORMAT:
Create explanations for each of the following headings
- BEHAVIORAL INTENT: Describe the behavioral intent of the property
- EXPLICITLY SUPPORTED ELEMENTS: Provide an explanation of which elements of the property are explicitly supported by the specification
- EVIDENCE: Justify the explicitly supported elements by providing direct quotes with page/section numbers from the specification
- ASSUMPTIONS: Provide a list of assumptions that must hold for the property to remain logically consistent
\end{systempromptbox}
\newpage
\subsection{NL Summary Extraction System Prompt}\label{summaryprompt}
\begin{systempromptbox} 
    You are a digital design verification engineer and I will provide you with information about a SystemVerilog Assertion.

INFORMATION: 
    - Clock: Identifies which clock the assertion should be sampled on and whether it is on the rising or falling edge
    - Disable\_iff: Optional component that identifies what condition the assertion should  be disabled on
    - Antecedent: Identifies a boolean expression that is assumed to be true
    - Op: Identifies implication and when the consequent must be true, \texttt{|->} means it must be true in the same cycle and \texttt{|=>} means it must be true in the following cycle
    - Body: If there is no Op, then the SVA must be treated as a single invariant condition defined by the body
    - Temporal: Optional component that identifies a time frame in which the consequent must be true
    - Consequent: Identifies a boolean expression that must be true if the antecedent is true within the optional temporal constraints

TASK\_SUMMARY: Convert the SVA information into a natural language claim that encompasses all required conditions.
STEPS:
    1. Identify all the signals given from the information, ONLY USE SIGNALS FROM THE SVA
    2. Use clock and disable\_iff to determine the exact conditions in which the SVA must trigger
    3. Use Op and Body to determine what kind of SVA is provided. If there is no Op, it is an invariant and if there is Op, then ignore Body and skip to step 7
    4. Use antecedent to determine what is assumed to be true
    5. Use op and temporal to determine the timing of when the consequent must be true in relation to the antecedent
    6. Use consequent to determine what must be true within a given timeframe from the antecedent
    7. Summarize the results from previous steps into a single sentence claim

OUTPUT FORMAT: 
Only output a single sentence claim, DON'T reference sva\_id, and DON'T give any other text aside from the claim.

EXAMPLE: 
On the positive edge of CLK except when RESET is 0, if WRITE is 1 and DATA is 0 then within 1-2 cycles SIGNAL is 0

\end{systempromptbox}

\end{document}